\documentclass[11pt,a4paper]{article}
\pdfoutput=1 

\usepackage{amsmath}

\usepackage{xspace}

\newcommand{\code}[1]{\texttt{#1}}

\newcommand{\latin}[1]{\textit{#1}\xspace}
\newcommand{\ie}{\latin{i.e.}}
\newcommand{\eg}{\latin{e.g.}}

\newcommand{\warning}[1]{{\PackageWarning{pltrdy_commands}{#1}}}

\usepackage{csquotes}
\MakeOuterQuote{"}

\usepackage[shortlabels]{enumitem}

\usepackage{amssymb}
\usepackage{pifont}
\usepackage{booktabs}
\usepackage{ctable} 

\usepackage{array}
\usepackage{makecell}
\usepackage{float}

\newcommand{\subhead}[2]{\thead{\textbf{#1} \\ #2}}

\usepackage{tabularx}
\newcolumntype{C}{>{\centering\arraybackslash}X}
\newcolumntype{L}{>{\raggedright\arraybackslash}X}
\newcolumntype{R}{>{\raggedleft\arraybackslash}X}

\usepackage{arydshln}

\usepackage{stackengine}

\usepackage{multirow}

\usepackage{rotating}

\newrobustcmd{\B}{\bfseries}

\newrobustcmd{\Bl}{}
\newrobustcmd{\Br}{}

\newrobustcmd{\MR}{{\color{red}\textbf{??}}}
\newrobustcmd{\TR}{{\color{blue}\textbf{!?}}}

\newrobustcmd{\extRes}[1]{\emph{#1}}
\newrobustcmd{\extResA}[1]{\extRes{#1\textsuperscript{*}}}
\newrobustcmd{\extResB}[1]{\extRes{#1\textsuperscript{$\star$}}}
\newrobustcmd{\extResC}[1]{\extRes{#1\textsuperscript{$\circ$}}}

\usepackage{graphicx}

\usepackage{float} 

\usepackage[export]{adjustbox}

\usepackage{caption}
\usepackage{subcaption}

\usepackage{subfloat}

\usepackage{bbm}

\newcommand{\card}[1]{\abs*{ #1 }}

\usepackage[hyperref]{ranlp2023}
\usepackage{breakurl}
\usepackage{times}
\usepackage{latexsym}

\usepackage{microtype}

\aclfinalcopy 

\title{LC-Score: Reference-less estimation of Text Comprehension Difficulty}

\author{Paul Tardy \\
  U31 \\ \url{https://u31.io} \\
  \texttt{pltrdy@gmail.com} \\\And
  Charlotte Roze \\
  U31 \\ \url{https://u31.io} \\
  \texttt{charlotte.roze@u31.io} \\\And
  Paul Poupet \\
  U31 \\ \url{https://u31.io} \\
  \texttt{paul.poupet@u31.io} \\}

\date{}

\newcommand{\langageclair}{Langage Clair\xspace}

\begin{document}   
\maketitle
\begin{abstract}
Being able to read and understand written text is critical in a digital era. However, studies shows that a large fraction of the population experiences comprehension issues. In this context, further initiatives in accessibility are required to improve the audience text comprehension.
However, writers are hardly assisted nor encouraged to produce easy-to-understand content. Moreover, Automatic Text Simplification (ATS) model development suffers from the lack of metric to accurately estimate comprehension difficulty
We present \textsc{LC-Score}, a simple approach for training text comprehension metric for any French text without reference \ie predicting how easy to understand a given text is on a $[0, 100]$ scale.
Our objective with this scale is to quantitatively capture the extend to which a text suits to the \textit{Langage Clair} (LC, \textit{Clear Language}) guidelines, a French initiative closely related to English Plain Language.
We explore two approaches: (i) using linguistically motivated indicators used to train statistical models, and (ii) neural learning directly from text leveraging pre-trained language models. We introduce a simple proxy task for comprehension difficulty training as a classification task.
To evaluate our models, we run two distinct human annotation experiments, and find that both approaches (indicator based and neural) outperforms commonly used readability and comprehension metrics such as FKGL.
\end{abstract}

   

\section{Introduction} \label{sec:introduction}
The ability to understand text is essential for a wide range of daily tasks. It enables individuals to stay informed, understand administrative forms, and have a full, unimpeded access to social and medical care.

Studies shows that a large fraction of the population experiences comprehension issues in their daily life. Almost half of the OECD population shows reading and written information comprehension difficulties \cite{OECD2013,Stajner2021}.

Such difficulties have a major impact in people's life. In France for example, the National Statistic Institute \cite{INSEE2012} reports that one person out of four has already abandoned an administrative procedure deemed too complicated to follow-along. 


In order to improve written text accessibility, initiatives such as Plain Language\footnote{\url{https://plainlanguagenetwork.org/plain-language/what-is-plain-language}} or \textit{Language Clair}  (LC, translates to \textit{Clear Language}) defines writing guidelines to produce clearer texts. 
Moreover, comprehension  makes its way into international standards and norms \cite{ISO24495,WCAG2018} but still lacks of concrete solution and measurable objectives.

With the rise of deep-learning approaches in natural language processing, as well as its recent successes in a wide variety of tasks (transcription, translation, summarization, question answering), Automatic Text Simplification is an interesting candidate for accessibility improvements at scale. However, system performances are difficult to measure due to the limitations of current automatic metrics \cite{Alva-Manchego2021:TheUnsuitability}. 

We hypothesize that the development of better text comprehension metrics could provide Automatic Text Simplification researchers with a way of validating their models while also to giving measurable objectives for the content editors to write clearer texts.

In this context, we focus our work in developing models for reference-less text comprehension evaluation as a scoring function for French texts \ie $s : \text{text} \mapsto \left[0, 100\right]$ reflecting how clearly written a text is.


In this paper, we present the following contributions:
\begin{itemize}[-]
    \item We introduce a simple approach to address comprehension evaluation as a classification task
    \item We introduce a set of linguistically motivated lexical, syntactic and structural indicators 
    \item We train both indicator based models and text-based Neural Models
    \item We evaluate our experiments thanks to two human annotation experiments using crowd sourced human judgement for one and expert rating for the second.
\end{itemize}

\section{Related Work} \label{sec:related_work}

Defining what makes a text difficult to understand is a complex task by itself. Multiple approaches are explored, like studying the age at which children acquires complex syntactic constructions in French \cite{Canut2014:Acquisition}; or relying on standardized foreign language levels such as the Common European Framework of Reference (CEFR), ranging from A1 to C2. \citet{Wilkens2022FABRA} uses this scale to study French as a Foreign Language difficulty.

In order to improve texts clarity, some organizations produced redaction guidelines \ie suggestions of good practices to write clear texts, such as Plain Language \cite{PLAIN2011} and, in French, \cite{ecrire-pour-etre-lu}. \citet{Gala2020:RecommandationsALECTOR} also published guidelines for adapting French texts to increase readability and comprehension. More closely related to our work, \citet{Francois2012} introduced a readability formula for French as a foreign language.

Automatic Text Simplification aims at generating simpler versions of a source texts. In literature, such models are usually evaluated using automatic metrics. Therefore, standard language level and redaction guidelines are hardly suitable to evaluate simplification models since it would require an expert judgement.
Automatic evaluation instead mostly rely on readability metrics such as FKGL \cite{Kincaid1975:FKGL}, SMOG \cite{Harry1969:SMOG} and Gunning fog Index \cite{Gunning1952:GunningFog}. Such metrics were designed with English in mind but can be used on French in practice. On the other hand, SAMSA \cite{Sulem2018:SAMSA}, a semantic metric, is currently not implemented for French, as discussed in  section \ref{sec:metrics}. 

Other approach include learning regression and classification models \cite{Martin2018ReferenceLess} or pre-trained language models \cite{Zhang2019:BERTScore}. However, \cite{Alva-Manchego2021:TheUnsuitability} found that automatic metrics remains unsuitable to evaluate progress in Automatic Text Simplification.

\section{Methods} \label{sec:methods}

\subsection{Baseline metrics} \label{sec:metrics}

In order to evaluate our work with respect to the literature we take the following existing readability metrics as baselines: FKGL \cite{Kincaid1975:FKGL}, SMOG \cite{Harry1969:SMOG}, Gunning Fog \cite{Gunning1952:GunningFog}.

The SAMSA metric \cite{Sulem2018:SAMSA} takes semantic into consideration. Even though it would be theoretically possible to adapt this metric for french, it is not yet implemented. We tried adapting existing implementation from EASSE \cite{Alva-Manchego2019EASSE} based on CoreNLP \cite{Manning2014CoreNLP} but it turned out to fail due to the lack of French lemmatization model. 







\subsection{Evaluate text comprehension difficulty as a classification task} \label{sec:classification_task}

Training a model to predict comprehension difficulty would require a text corpus annotated with comprehension scores. However, to the best of our knowledge, their is no such corpus for the general audience and of sufficient size to envision model training.
In this context, we suggest to rely on a simpler proxy task consisting of a classification between \textit{simple} and \textit{complex} texts. Defining what makes a text simple or complex here is difficult. In order to bypass this question, we uses pairs of content sources such as one is roughly a simplified version of the other:

\paragraph{Encyclopedia articles} based on French Wikipedia (\textit{complex}) and its simpler alternative, Vikidia (\textit{simple}), designed for 8-13 years old readers. We only took into consideration the introduction paragraph as it is a concise and synthetic presentation of the article. Articles are aligned \ie the corpus consists in $(simple, complex)$ pairs. 

\paragraph{International Radio Journal Transcriptions} with 
    \textit{France Culture international press review} (\textit{complex})
        \footnote{
            \small{\url{https://www.radiofrance.fr/franceculture/podcasts/revue-de-presse-internationale}}
        } 
    and 
    \textit{RFI Journal En Français Facile} (\textit{simple}),
        \footnote{
            \small{\url{https://francaisfacile.rfi.fr/fr/podcasts/journal-en-fran\%C3\%A7ais-facile/}}
        } 
    aimed at french speakers that do not speak the language on a daily basis. Articles have similar subjects (international news) but are not aligned strictly speaking \ie there is no $(complex, simple)$ pairs for a given article.
    We report statistics about this new corpus in table \ref{tab:datasets}.


\begin{table}
    \begin{center}

    \setlength{\tabcolsep}{6pt}
    \begin{tabular}{l*4{c }} 
    
    \toprule
        \textbf{Corpus}
            & \textbf{\#T}
            & \textbf{\#W/\#T}
            & \textbf{\#W/\#S}
        \\\midrule
        Wikipedia & 25812 & 144 & 26.0
        \\ Vikidia & 25812 & 80 & 18.9
       \\\midrule
         France Culture & 1402 & 1106 & 28.8
        \\  \begin{tabular}{@{}c@{}}Journal en \\ Français Facile\end{tabular} & 1555 & 1494 & 19.0
    \\\bottomrule
    \end{tabular}
        \caption{Comprehension Classification Datasets: number of texts per corpus ($\#T$), average word per text ($\#W/\#T$) and average word per sentence $\#W/\#S).$\label{tab:datasets}}
    \end{center}
\end{table}

\subsection{Linguistic Indicators} \label{sec:indicator_models}

Deriving from works on \langageclair we introduce a set of complexity indicators.
Indicators varies from lexical difficulties (\ie a word difficulty score) to syntactic difficulties or sentences parse tree height.
Indicators are detailed below.

Indicators are detected based on our own rules implementation using SpaCy pipeline based on both dependency and constituency
parsing respectively using \code{fr-dep-news-trf}\footnote{\url{https://spacy.io/models/fr\#fr_dep_news_trf}} and \code{benepar}\footnote{\url{https://github.com/nikitakit/self-attentive-parser}}.

\paragraph{Lexical Indicators (5)} These are indicators of difficulties at word level. We use a word difficulty score based on word frequencies in corpora of different difficulty levels: elementary school textbooks of various grades from Manulex \cite{lete:hal-00733549} and French as a Foreign Language textbooks of various CEFR (Common European Framework of Reference for Languages) levels from FLELex \cite{francois-etal-2014-flelex}. Lexical indicators also include abbreviations, acronyms, named entities and numerical expressions.

\paragraph{Sentence Length Indicators (3)} We measure sentences lengths with averages of words per sentence; dependency and constituency tree heights.

\paragraph{Syntactic Indicators (17)} Several difficulties on the syntactic level in sentences are identified, which are related to sentence structure: coordinate clauses, relative clauses, adverbial clauses, participle clauses, cleft structures, interpolated clauses, appositive phrases, enumerations, etc.). Information about verb forms are also detected: non-finite clauses, passive voice, complex verbal tenses, conditional mood. Negations marks, complex noun phrases and text spans between brackets are also included in syntactic indicators. 

\paragraph{Structure Indicators (3)} Two indicators are related to the presence of connectives and their potential complexity, estimated by syntactic information (\eg clause position for conjunction connectives, sentence initial position for adverbial connectives) and information from a French connectives lexicon \cite{roze:inria-00511615}. A third indicator counts temporal breaks (\ie a tense change) within text paragraphs.

We train models using \texttt{sklearn}: two linear models (Linear SVC and Ridge) for fairer comparison to linear readability metrics, and 2 non-linear (Random Forest and Multi Layer Perceptron)

 

\subsection{Neural Methods based on Text} \label{sec:neural_models}
Even though indicator-based approaches rely on linguistic motivations, they lack the possibility to learn from deeper relationships throughout the text such as the subject, the context and the semantic that might carry essential information to infer comprehension difficulty. This is the reason why we chose to compare indicator-based methods with deep learning approaches directly relying on text.

We use two French pre-trained language models such as BARThez \cite{Eddine2020:BARThez} and CamemBERT \cite{Martin2020Camembert} fine-tuned with a classification (C) or a regression objective (R).

\section{Comprehension Difficulty Annotation} \label{sec:evaluation}
We ran two human annotation experiments in two different contexts: the first one using Mechanical Turk, a crowd-sourcing platform to receive annotations of French speakers from general audience (\ref{sec:eval_mturk}); the second based on the feedback of \langageclair  experts in our team (\ref{sec:eval_experts}).

\subsection{Crowd-sourced Human Annotation} \label{sec:eval_mturk}

In order to get the most reliable annotations we follow \cite{Kiritchenko2017BestWorstScaling} and use a Best-Worst Scaling (BWS) technique. They recommend to use comparison task instead of direct assessment \ie directly giving a note to a given text. More specifically, BWS compares $k$ (typically $k=4$) simultaneous examples and asks the annotator to select the best one and the worst one with respect to the dimension of interest  (text comprehension difficulty in our context). 

When annotating texts of up to $200$ words, preliminary experiments showed us that comparing $k=4$ simultaneous texts was too long and fastidious.
In this light, we reduce to $k=3$.

The annotation counts $T=48$ news articles (up to 200 words). Each text is present in $e=12$ different examples of $k=3$ texts. Examples are annotated by $a=3$ separate annotators in a total of $26$.
We end up with a total of $E = (T \times e) / k = 192$ examples, and $E \times a $ annotation \ie for any three texts $\left\{ T_a; T_b; T_c \right\} $ the annotation task consist in submitting an ordered set \eg $T_c > T_a > T_b$.

Each text $T_i$ is associated with an annotation score by $score(i) = \#best\%(i) - \#worst\%(i)$ with $\#best\%(i)$ (resp. $\#worst\%(i)$) representing the frequency at which $T_i$ was evaluated the best (resp. worst) text out of the 3.

In order to measure the reliability of an annotation experiment, a common practice is to measure inter-annotation agreement. However, in a BWS process, each annotators is presented with a different set of examples which makes the concept of annotator agreement less relevant. Moreover, disagreement is even beneficial to produce accurate annotation: for two items $A$ and $B$ of similar difficulty, we can expect half of the annotator to rate $A > B$ and the other half $B > A$. From this apparent disagreement emerges diversity that actually reinforce score accuracy.
For this reason, BWS is instead evaluated in terms of reproductibility metrics like Split Half Reliability (SHR). SHR is the correlation between two randomly sampled half of the annotation. In practice, we average SHR over $1000$ iterations to rule out randomness.

\subsection{Expert Annotation} \label{sec:eval_experts}
In addition to crowd-sourced corpus, our team built a small corpus of $74$ texts annotated with difficulty scores.
We selected $37$ texts originating from news articles, literature, and customer support mails. In addition, we provide $37$ manually simplified versions following \langageclair methodology. Each of the $74$ resulting texts were then scored on a $\left[0, 100\right]$ scale by $4$ LC experts from our team.

To make sure we obtained good quality annotation, we measure annotator agreement with Intra-class Correlation Coefficient (ICC2, \citealp{Shrout1979:ICC:Intraclasscorrelations}). ICC2 ranges from $0$ (no agreement) to $1$ (perfect agreeement). 


\section{Results} \label{sec:results}

\subsection{Annotation results} \label{sec:results_annotation}
Annotations experiments text length metrics and reliability measure are reported in table \ref{tab:human_eval_experiments}.

\paragraph{Good reliability from MTurk and Expert} even though our annotation experiments are very different in terms of annotators and process, both shows high reliability measures achieving respectively an SHR correlation of $64.7$ (MTurk) and an Intraclass Correlation Coefficient of $74.6$ (Experts).


\paragraph{Filtering MTurk workers does not increase reliability} A common practice when involving crowd-sourced annotation is to filter-out users that shows the lowest agreement. Even though we discussed in \ref{sec:eval_experts} that agreement is not considered to be the most relevant metric for BWS annotation, we challenge this hypothesis by calculating worker agreement rate based on how often a given user submits the same result than another worker. Then, we suppose that workers with the lowest agreement rate might add noise to the experiment so we might want to exclude them. However, results showed the opposite: filtering out workers does not increase reliability in terms of SHR, no matter the agreement rate of each. This observation is in line with the hypothesis that annotator disagreement is expected and beneficial in a BWS annotation experiment.

\begin{table}[h]
    \begin{center}

    \setlength{\tabcolsep}{6pt}
    \begin{tabular}{l*4{c }} 
    \toprule
        
            & \textbf{MTurk}
            & \textbf{Expert}

        \\\midrule
        \#T & 48 & 37 / 37   \\
        \#W/\#T & 183 & 190 / 209   \\
        \#W/\#S & 25 & 28 / 13   \\ \midrule
        \#Annotators & 26 & 4   \\
        Type & BWS & RS \\
        Reliability Measure & SHR & ICC2   \\
        Reliability & 64.7 & 74.6	 \\
    \bottomrule
    \end{tabular}

    \caption{Human Annotation Experiments. Corpus are reported with number of texts per corpus ($\#T$), average word per text ($\#W/\#T$) and average word per sentence $\#W/\#S)$. Since Expert is aligned, metrics are reported for both sides. Experiments uses two different annotation processes (i) Best Worst Scaling (BWS) evaluated in term of Split Half Reliability (SHR) and (ii) Rating Scale in $[0, 100]$ (RS, 100 is best) evaluated with Intraclass Correlation Coefficient (ICC2). }
    \label{tab:human_eval_experiments}
    
    \end{center}
    \end{table}

\subsection{Scoring results} \label{sec:results_scoring}
First, we evaluate model performances with respect to their own training by measure accuracy on their validation set: a $10\%$ held-out subset from the training set. Validation accuracy is used to select the best hyper-parameters and training iterations for each models.

Models are then evaluated against human annotations from MTurk and Experts using Spearman Rank Correlations ($\rho$).

Results are reported in \autoref{tab:results_correlations}.
Our approaches show better correlations with the human judgement than readability metrics.
Models trained from indicators achieves the highest correlations, with Random Forest being the best on both evaluation sets, MTurk and Expert.

It is also interesting that even simple linear statistical models based on our indicators outperforms readability metrics therefore arguing in favor of this indicator set. In particular, the Ridge Regression model outperform FKGL by $14.76$ and $10.55$ correlation point respectively on MTurk and Expert.

Readability metrics seems complementary in that FKGL achieve better correlation on MTurk evaluation while Gunning Fog does on Expert.

Similarly, we observe sensible differences between Camembert training objectives, with the regression (R) being better on MTurk and classification (C) on Expert.


\begin{table}
    \begin{center}
    \setlength{\tabcolsep}{6pt}
    \begin{tabular}{l*4{c }} 
    
    \toprule
        \subhead{Model}{\hfill}
            & \subhead{Valid}{$acc\%$ }
            & \subhead{MTurk}{$\rho$ }
            & \subhead{Expert}{$\rho$ }
        \\\midrule
        SMOG  & - & -18.68 & -73.09 \\
        Gunning Fog  & - & -12.59 & \textbf{-82.14} \\
        FKGL  & - & \textbf{-19.66} & -77.54 \\
        \midrule
        Linear SVC  & 73.07 & 20.94 & 69.37 \\
        Ridge & - & 27.58 & 86.44 \\
        MLP  & 75.31 & 32.56 & 85.73 \\
        Random Forest  & \textbf{77.20} & \textbf{34.42} & \textbf{88.09} \\
        \midrule
        BARThez  & 79.64 & 23.16 & 58.41 \\
        Camembert(R)  & \textbf{91.01} & \textbf{28.35} & 75.85 \\
        Camembert(C)  & 90.15 & 18.44 & \textbf{84.73} \\

    \bottomrule
    \end{tabular}

    \caption{Scoring models Spearman correlations ($\rho$) with human judgement. (C) and (R) respectively indicates classification and regression training objective.}
    \label{tab:results_correlations}
    
    \end{center}
\end{table}



\section{Discussions} \label{sec:Discussions}

Results shows a large improvement of human judgement correlation in favor to our approaches over existing readability metrics. Moreover, indicator based method outperform neural models fine-tuned from pre-trained model. Neural models' results are promising and could be extended with longer training time and adapting their training objective to produce equally distributed scores. 

In addition to outperforming neural models, indicator based model are far cheaper to train and predict with since they does not require GPU.
Being indicator-based makes it easier to interpret and more predictable than neural models, and thus might deliver a better user experience.
We observed Neural models we trained tend to produce very polarized output probabilities \ie either very close to $0$ or to $1$. That's not a problem to quantitatively evaluate the resulting score, but it should probably be adapted to output equally distributed scores in order to be more intuitive.





\section{Conclusion} \label{sec:conclusion}
Developing methods to accurately measure written text comprehension difficulty is a key challenge that would help better assessing the quality of Automatic Text Simplification models, and provide with a tool for editors to produce texts that are simpler to understand.

We explore multiple approaches for training a reference-less metric based on a simple classification task.
Our systems rely either on linguistic indicators or directly from text. 

To evaluate our models, we two human annotation experiments. The first involves crowd-sourced workers, asked to compare text based on their comprehension difficulties using Best Worst Scaling with $k=3$. In the second experiment, texts are simplified then rated on a $[0, 100]$ scale by experts from our team.

Both neural and indicator based methods shows promising results and largely outperform other broadly used readability metrics, on both crowd-sourced and expert human annotations. Even simple linear models largely outperform readability metrics which adds an evidence against using it to estimate text comprehension complexity.



As further researches, we suggest exploring multi-lingual neural training. This would have the obvious benefit of overcoming the language restriction of our work while also mutualizing learning from each language and unifying comprehension difficulties estimation accross languages.



\section{Lay Summary}
Nowadays, most services use the Internet as their primary way of communicating.
Therefore, being able to read and understand texts is really important.
But a lot of people have difficulties reading and understanding so it is not simple for them to access information or complete administrative procedures.

We introduce a method to calculate a difficulty score for French texts. A score of 0 means that the text is really difficult to understand, whereas a score of 100 means it is really clear. 
We suggest that developing such a score is a first step toward helping people write easier texts.
We gathered two categories of texts: some that we consider easy to understand and others that we consider difficult to understand. Then, we trained models to predict whether a text is categorized as "easy" or not. After training, we use the predictions as our scoring method: the score corresponds to the probability (multiplied by 100) that a text is categorized as easy by the model.

We explored two kinds of models. For the first one, we count different kinds of linguistic difficulties and give them to the model to predict the difficulty.
The second kind of model is deep neural networks that have already been trained to learn French. We specialize it in predicting the difficulty based on the text by providing examples of texts and their difficulties.

To measure how relevant our models are, we asked people on the Internet as well as experts to give their opinions on texts. In particular, they were given texts and should determine how difficult they are.
We found that people agreed more with our method's scores than with other existing scoring methods.


\bibliographystyle{acl_natbib}
\bibliography{mendeley,extra_bib}


\end{document}